\documentclass{article}
\usepackage{ijcai25}

\usepackage{times}
\usepackage{soul}
\usepackage{url}
\usepackage[hidelinks]{hyperref}
\usepackage[utf8]{inputenc}
\usepackage[small]{caption}
\usepackage{graphicx}
\usepackage{amsmath}
\usepackage{amsthm}
\usepackage{booktabs}
\usepackage{algorithm}
\usepackage{algorithmic}
\usepackage[switch]{lineno}
\usepackage{subcaption}
\usepackage{makecell}
\usepackage{multirow}
\title{Contrastive Domain Generalization for Cross-Instrument Molecular Identification in Mass Spectrometry}
\author{
Seunghyun Yoo$^1$\thanks{These first co-authors contributed equally to this work.}\and
Sanghong Kim$^1$\footnotemark[1]\And
Namkyung Yoon$^1$\footnotemark[1]\and
Hwangnam Kim$^1$\thanks{Corresponding author.}\\
\affiliations
$^1$School of Electrical Engineering, Korea University, Seoul 02841, Korea\\
\emails
\{seunghyunyoo, sanghongkim, nkyoon93, hnkim\}@korea.ac.kr
}

\begin{document}

\maketitle

\begin{abstract}
Identifying molecules from mass spectrometry (MS) data remains a fundamental challenge due to the semantic gap between physical spectral peaks and underlying chemical structures. Existing deep learning approaches often treat spectral matching as a closed-set recognition task, limiting their ability to generalize to unseen molecular scaffolds. To overcome this limitation, we propose a cross-modal alignment framework that directly maps mass spectra into the chemically meaningful molecular structure embedding space of a pretrained chemical language model. 
On a strict scaffold-disjoint benchmark, our model achieves a Top-1 accuracy of 42.2\% in fixed 256-way zero-shot retrieval and demonstrates strong generalization under a global retrieval setting.
Moreover, the learned embedding space demonstrates strong chemical coherence, reaching 95.4\% accuracy in 5-way 5-shot molecular re-identification. These results suggest that explicitly integrating physical spectral resolution with molecular structure embedding is key to solving the generalization bottleneck in molecular identification from MS data.
\end{abstract}

\section{Introduction}\label{sec1}

Mass spectrometry (MS) is one of the most widely used analytical techniques for chemical identification in scientific, industrial, and forensic fields \cite{wood2006recent}. 

However, the spectrum measured in heterogeneous equipment can often show significant variations in the proficiency of the measurer, peak resolution, noise characteristics, and fragmentation patterns.
These equipment-dependent discrepancies lead to severe distribution shifts, which adversely affect the identification and detection performance of existing deep learning models when measuring devices or molecular scaffolds differ from the training distribution, limiting their immediate use on unseen molecules or new equipment settings.


While recent advances in cross-modal learning and chemical language models have improved the semantic understanding of molecular structures, aligning MS spectra with chemically meaningful molecular structure embeddings remains a major challenge.
The MS spectral peaks depend strongly on the collision energy of the chemicals and the detector configuration. Molecular structure embeddings capture structural semantics, but they do not directly model the physicochemical behavior manifested in mass spectrometry signals.
Moreover, publicly available MS datasets, such as MassBank, contain very heterogeneous measurement domains, creating a significant semantic gap between the physical spectrum and molecular representations \cite{horai2010massbank}.

To address these limitations, we propose a contrastive cross-modal transfer learning framework that aligns mass spectra from heterogeneous instruments in a molecular structure embedding space to learn domain-invariant molecular representations. This approach is based on two observations.

First, cross-modal constrastive alignment allows molecular structure embeddings to inherit chemically meaningful structures from pre-trained molecular language models, thereby enhancing generalization to unseen chemical compounds.
Additionally, the proposed alignment framework naturally extends to few-shot scenarios, as chemically consistent embeddings facilitate reliable similarity-based classification even with limited support samples.

The contributions of this work are summarized as follows:
\begin{itemize}
\item We reformulate molecular identification from mass spectrometry as a cross-modal semantic alignment problem and introduce a contrastive dual-encoder framework that learns to align spectral representations with molecular structure embeddings, thereby achieving zero-shot generalization to unseen chemical scaffolds.

\item Instead of combining heterogeneous modalities, we employ a chemical language model whose embeddings capture molecular structural semantics, allowing spectral signal patterns to align directly with molecular structure embeddings and narrowing the semantic gap between the two modalities.

\item We introduce an MS spectral encoder incorporating mass-to-charge transformation, intensity normalization, and Gaussian Fourier projections, and demonstrate their effectiveness through extensive zero-shot and few-shot evaluations in rigorous scaffold-separation settings, achieving improved accuracy and reduced episode variance among heterogeneous devices.
\end{itemize}

The rest of this paper is organized as follows. Section 2 provides a related study on MS-based molecular data of chemistry and transfer learning. Section 3 describes the proposed framework. Section 4 presents the experiments and evaluations. Section 5 discusses the conclusions.


\section{Related Work}\label{sec2}
In this section, we review previous studies related to MS-based molecular identification, transfer learning, and few shot learning.
We focus on existing approaches dealing with spectral representation learning and cross-domain generalization, and discuss their limitations in heterogeneous instrumental conditions.
With this review, we clarify the problem we are trying to solve as a cross-modal semantic alignment problem rather than a simple pattern recognition task.
\subsection{Mass spectrometry molecular identification}\label{sec2.1}

MS is one of the most widely used analytical techniques for molecular identification, as it captures characteristic fragmentation patterns that reflect underlying chemical structures.
Recent studies have applied deep learning models to MS data for molecular identification, spectral matching, and hazardous substance detection
\cite{yoon2024unveiling,yoon2024pioneering,schymanski2014identifying,chen2024screening}.

Most existing approaches treat MS-based identification as a closed-set classification or retrieval problem, relying on supervised learning under homogeneous measurement conditions \cite{scott1988classification,xu2025large}.
Although convolutional and recurrent neural networks have been widely adopted for spectral representation learning, existing methods remain fundamentally constrained in cross-instrument generalization due to device-specific peak distortions, collision-energy–dependent fragmentation, and resolution variability.

In contrast, our work aims to address molecular identification under device-induced distribution shifts in MS data, reframing the task as cross-modal semantic alignment rather than device-specific spectral pattern learning.

By explicitly modeling fine-grained mass resolution and aligning spectral representations with molecular structure embeddings, the proposed framework addresses domain shifts that arise from both molecular diversity and instrumental variability.

\subsection{Transfer learning}\label{sec2.2}

Transfer learning is being studied as a key strategy for improving generalization in tasks where the learning domain and the target domain are different.
Traditional domain adaptation methods have been done by aligning the feature space between source and target data to reduce distribution mismatch. Recent approaches use metric learning, contrast objective, or cross-modal alignment to learn transfer-friendly embeddings. These methods have been applied in various fields, including image recognition, speech processing, and molecular representation learning \cite{wei2020universal,kim2022transferring}.

In chemical analysis, transfer learning has been applied to tasks such as molecular property prediction and cross-domain spectral matching. However, most existing approaches assume uniform measurement conditions or rely on closed supervised settings with labeled target-domain samples.

Unlike previous studies, our work specifically focuses on detecting substances measured across heterogeneous instruments in chemical analysis. To address this challenge, we propose a cross-modal alignment framework that bridges MS spectra and molecular structure embeddings while jointly learning domain-specific spectral representations. This aims at generalization to unseen molecules and previously unobserved measuring instruments, addressing important limitations in existing MS-based identification systems.

\begin{figure}
\centering \makeatletter\IfFileExists{./figures/overall4.png}{\includegraphics[width=0.45\textwidth]{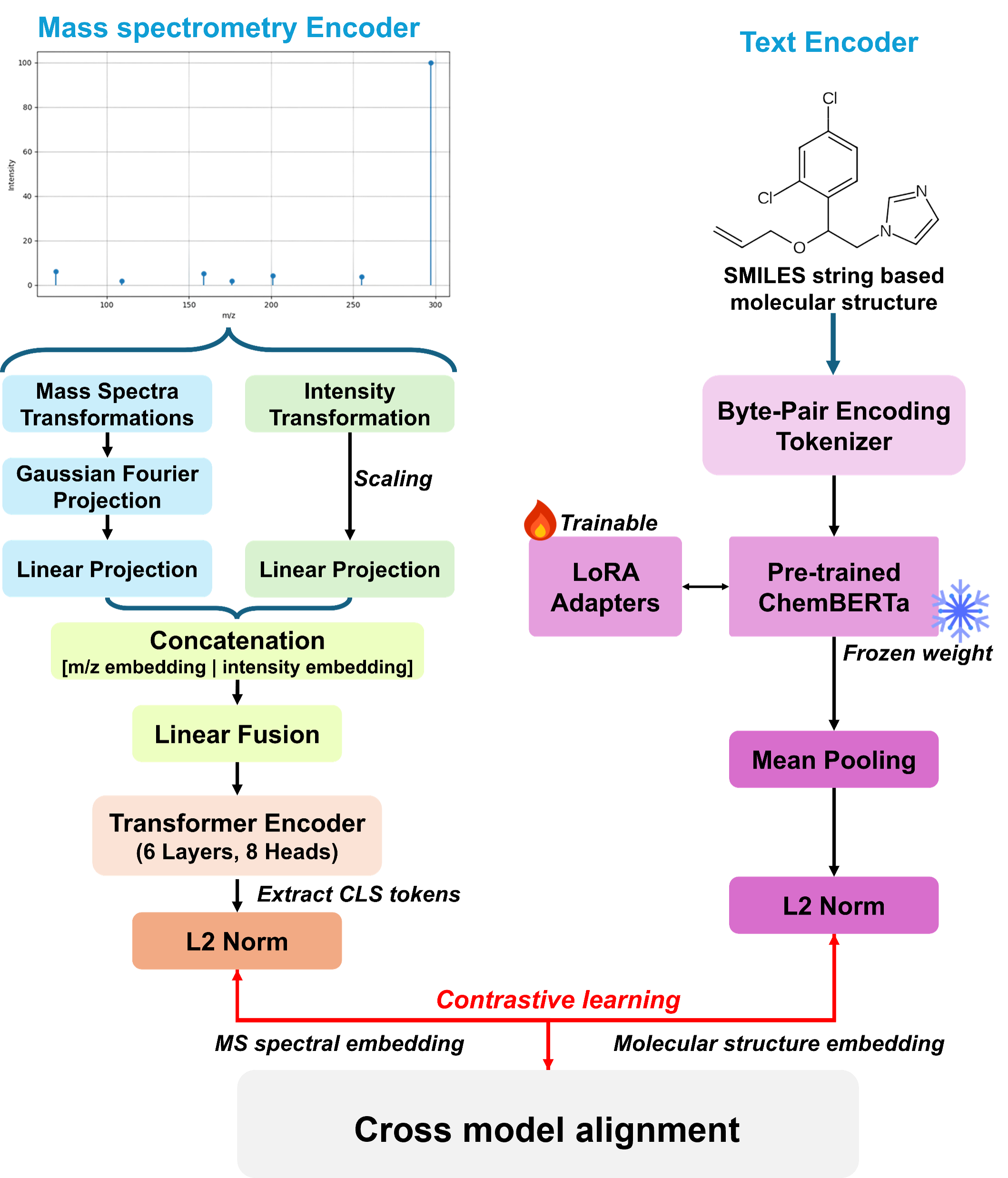}}{}
\makeatother 
\caption{{Overall structure of the proposed framework.}}
\label{overall}
\end{figure}
\subsection{Few-shot learning}\label{sec2.3}
Few-shot learning is an artificial intelligence technique that aims to recognize new classes using only a small number of labeled examples. 

The general approach relies on episode learning, where the model learns to compare support samples with query samples using metric-based or embedding-based strategies \cite{laenen2021episodes,qiao2019transductive}.
These methods have been studied to be effective in areas where annotated data are limited or new classes frequently appear.
However, existing learning methods implicitly assume homogeneous feature spaces between tasks, which is problematic when measuring spectra on heterogeneous instruments.

In the proposed work, we use few-shot learning to evaluate whether the learned spectral representation is generalized to new chemical classes and previously unseen instruments. By incorporating cross-modal alignment and domain-specific encoding, the proposed framework enables more stable similarity estimation under heterogeneous sensing conditions, improving performance in both zero-shot and few-shot molecular identification.

\section{System Design}\label{sec:system_design}
This section describes the overall design of the proposed framework.
Our framework is built on a dual encoder architecture that jointly models MS signals and molecular structure embedding space.
As shown in Fig.~\ref{overall}, the main design objective is to learn domain-invariant and molecular structure embeddings that generalize across heterogeneous instruments and previously unseen molecular scaffoldings, such as instrument-specific spectral distortions, acquisition condition variability, and molecular structural diversity.
To this end, the framework is designed to explicitly address both instrument-induced variability and molecular structural diversity.

\subsection{Dataset Curation and Partitioning}
The MS data utilized for training is from MassBank \cite{horai2010massbank}. To ensure rigorous evaluation, we sanitize the raw data by parsing all SMILES strings with RDKit \cite{rdkit}, discarding entries with invalid syntax or ambiguous connectivity.

We perform a \textbf{strict scaffold-disjoint split} based on the InChIKey. The InChIKey consists of three blocks, where the first block uniquely encodes the molecular skeleton, and the rest of the blocks represent stereochemistry and ionization-related variations that do not alter the core scaffold \cite{heller2015inchi}. By splitting the dataset based on this first block, we ensure that stereoisomers, which typically exhibit almost identical mass spectra, are not distributed across training and test sets. This scaffold-disjoint protocol enforces a realistic evaluation setting in which the model must infer chemical identity from molecular structure embedding principles rather than memorizing spectral fingerprints.

\subsection{Mass Spectra Encoder}
Raw mass spectra consist of sparse peaks defined by mass-to-charge ratios $m/z$ and intensities $I$. We apply a two-stage transformation pipeline to make this physical signal amenable to deep learning.

The effectiveness of this preprocessing is further analyzed in ablation studies, demonstrating that it contributes to both accuracy and stability.

\subsubsection{Mass Domain Transformation}
Raw mass spectra exhibit non-uniform measurement precision, where the peak width scales linearly with the mass-to-charge ratio ($m/z$) \cite{gross2017mass}.

From a machine learning perspective, this introduces heteroscedasticity in the input features. The significance of a numerical difference $\Delta m$ is not constant; high-mass regions exhibit broader spectral features compared to low-mass regions. This hinders ability of the model to learn scale-invariant representations, as the embedding space must otherwise account for varying feature density across the mass range.

To stabilize this variance and compress the dynamic range, we apply a logarithmic transformation:
\begin{align} 
    x_{mz} = \ln(m/z + 1). 
\end{align}
This transformation converts the instrument's constant relative peak width into a constant absolute peak width in the transformed space $x_{mz}$, ensuring that the input resolution remains consistent for the network's spectral encoder.

\subsubsection{Intensity Transformation}
Spectral intensities follow a heavy-tailed power-law distribution, where a single base peak often dominates hundreds of low-abundance diagnostic ions \cite{stein1994optimization}. Raw intensity values often span several orders of magnitude, causing numerical instability in neural networks. The loss function becomes exclusively driven by the base peak, causing the model to ignore trace signals, while large absolute values lead to exploding gradients and NaN loss values during backpropagation. To compress this dynamic range while preserving relative peak ordering, we apply Root Mean Normalization:

\begin{align}I' = \frac{\sqrt{I}}{\max(\sqrt{I})}.\end{align}
Unlike logarithmic scaling, the square root transformation provides a balanced gradient flow. It enhances the visibility of low-intensity isotopic satellites without completely flattening the prominence of major fragments. The division by $\max(\sqrt{I})$ ensures all inputs lie within $[0, 1]$, guaranteeing numerical stability. Finally, each spectrum is represented as a sequence $S = \{(x_{mz}^{(i)}, I'^{(i)})\}_{i=1}^N$, sorted by transformed mass.

\subsubsection{Gaussian Fourier Projection}
We project the scalar mass $x_{mz}$ into a high-dimensional Fourier space using a fixed Gaussian matrix $\mathbf{B} \sim \mathcal{N}(0, \sigma^2)$ \cite{tancik2020fourier}:
\begin{align} 
    \gamma(x_{mz}) = [\cos(2\pi \mathbf{B} x_{mz}), \sin(2\pi \mathbf{B} x_{mz})]. 
\end{align}
We set $\sigma=30$ as a fixed hyperparameter to provide sufficiently rich high-frequency mass features in the Fourier space, which helps preserve fine-grained peak variations that can be damped by spectral bias in deep networks \cite{rahaman2019spectral}.
Compared to standard positional encodings or learned embeddings, Gaussian Fourier projection explicitly preserves high-frequency mass variations that are otherwise suppressed by spectral bias.

By explicitly encoding fine-grained mass defects, the proposed projection allows the model to distinguish chemically meaningful high-frequency variations from incidental instrument-induced noise, resulting in a more stable embedding space.

\subsubsection{Orthogonal Channel Fusion}
To prevent the entanglement of chemical identity (m/z) and abundance (intensity), we treat the m/z–intensity pair as a single MS observation and encode them through orthogonal channels before linear fusion:
\begin{align} 
    \mathbf{e}_i = \text{Linear}( \gamma(x_{mz}) \mathbin\Vert \text{MLP}(I') ). 
\end{align}
This ensures that the attention mechanism can independently attend to where a peak is and how strong it is. The sequence is processed by a Transformer Encoder \cite{vaswani2017attention} which consists of 6 layers, 8 heads with a learnable CLS token to aggregate the global context of the entire fragmentation pattern into a single vector representation.

\subsection{Molecular Structure Encoder}
We utilize ChemBERTa \cite{chithrananda2020chemberta} to extract semantic embeddings from SMILES strings, leveraging its pre-trained knowledge of chemical grammar \cite{weininger1988smiles}.

\subsubsection{Low-Rank Adaptation (LoRA)}
To align the chemical space with the spectral domain, we apply LoRA \cite{hu2021lora} to the query $\mathbf{W}_q$, key $\mathbf{W}_k$, and value $\mathbf{W}_v$ projections:
\begin{align} 
    \mathbf{W} = \mathbf{W}_0 + \mathbf{B}\mathbf{A}. 
\end{align}
Unlike standard implementations, extending adaptation to the key projection introduces the expressivity required to bridge the significant modality gap using less than 1\% of trainable parameters.

\subsection{Cross-Modal Contrastive Alignment}
We align the spectral embedding $\mathbf{z}_s$, derived from MS spectra represented as unordered m/z-intensity pairs, and the molecular structure embedding $\mathbf{z}_m$ using the InfoNCE loss~\cite{oord2018representation}:
\begin{align} 
    \mathcal{L} = - \frac{1}{N} \sum_{i=1}^{N}
    \log \frac{\exp(\text{sim}(\mathbf{z}_{s,i}, \mathbf{z}_{m,i}) / \tau)}
    {\sum_{j=1}^{N} \exp(\text{sim}(\mathbf{z}_{s,i}, \mathbf{z}_{m,j}) / \tau)},
\end{align}
where $N$ denotes the batch size.

We use cosine similarity with $\ell_2$-normalized embeddings and set the temperature $\tau=0.07$ following common dual-encoder retrieval practice.
This asymmetric spectrum-molecular formulation reflects the work of downstream molecular search.

This contrastive objective encourages a chemically meaningful and globally consistent molecular structure embedding space, reducing embedding drift and sensitivity to episodic sampling during few-shot evaluation.

\subsection{Overall Framework Summary}
The proposed framework operates as a unified end-to-end dual-encoder pipeline. By integrating preprocessing, high-resolution spectral encoding, and cross-modal alignment, the system achieves robustness against acquisition variability and generalizes effectively to novel molecular scaffolds.

\begin{algorithm}[t]
    \caption{Heterogeneous modality Contrastive Training}
    \label{training}
    \begin{algorithmic}[1]
        \REQUIRE Dataset $\mathcal{D} = \{(S_i, M_i)\}_{i=1}^N$ (Spectra, SMILES)
        \REQUIRE Parameters $\sigma=30, \tau=0.07$
        \ENSURE Encoders $f_\theta, g_\phi$ \\
        \textbf{Preparation:} 
        \STATE $\mathcal{D}_{\text{train}}, \mathcal{D}_{\text{test}} \leftarrow$ \textsc{ScaffoldSplit}($\mathcal{D}$) \\
        \textbf{Preprocessing:}
        \FORALL{$S \in \mathcal{D}_{\text{train}}$}
            \STATE Sample $m, I$ from $S$
           \STATE $x_{mz} \leftarrow \ln(m + 1)$
            \STATE $I' \leftarrow \sqrt{I} \,/\, \max(\sqrt{I})$
        \ENDFOR \\
         \textbf{Initialization:}
         \STATE  $\mathbf{B} \sim \mathcal{N}(0, \sigma^2)$, $g_\phi \leftarrow$ ChemBERTa+LoRA \\
        \textbf{Training Loop:}
        \WHILE{not converged}
            \FOR{each batch $(S, M)$ in $\mathcal{D}_{\text{train}}$}
                \STATE $S \leftarrow \text{Sort}(S, \text{ascending})$
                \STATE $\gamma(x_{mz}) \leftarrow [\cos(2\pi \mathbf{B} x_{mz}), \sin(2\pi \mathbf{B} x_{mz})]$
                \STATE $\mathbf{h}_{emb} \leftarrow \text{Linear}(\gamma \mathbin\Vert I')$
                \STATE $\mathbf{z}_s \leftarrow f_\theta([\texttt{CLS}] \oplus \mathbf{h}_{emb})$
                
                \STATE $\mathbf{z}_m \leftarrow g_\phi(M)$
                
                \STATE $\mathbf{z}_s, \mathbf{z}_m \leftarrow \text{Normalize}(\mathbf{z}_s), \text{Normalize}(\mathbf{z}_m)$
                \STATE $\mathcal{L} \leftarrow \text{InfoNCE}(\mathbf{z}_s, \mathbf{z}_m; \tau)$
                \STATE Update $\theta, \phi$ to minimize $\mathcal{L}$
            \ENDFOR
        \ENDWHILE
    \end{algorithmic}
\end{algorithm}

\section{Experiments}\label{sec:experiments}
In this section, we evaluate the proposed framework in a challenging and realistic setting designed to evaluate its generalization capabilities.
The proposed framework is implemented as Algorithm~\ref{training} to analyze episode variance in addition to average accuracy to evaluate robustness and stability in different evaluation episodes.
\subsection{Experimental Setup}
\subsubsection{Dataset}
We utilized the MassBank dataset, processed via the \textbf{strict scaffold-disjoint splitting protocol} detailed in Section \ref{sec:system_design}. The resulting splits are as follows:
\begin{itemize}
    \item \textbf{Training Set:} Consists of 106,189 spectra corresponding to 14,093 unique scaffolds.
    \item \textbf{Test Set:} Consists of 26,159 spectra corresponding to 3,524 novel scaffolds that are structurally disjoint from the training data.
\end{itemize}

\begin{table*}[t]
\centering
\small
\resizebox{0.88\textwidth}{!}{
\begin{tabular}{llllc}
\toprule
\textbf{Evaluation Setting}
& \textbf{Candidate Pool}
& \textbf{Pool Size}
& \textbf{Method}
& \textbf{Hit@1 (\%)} \\
\midrule

\multirow{7}{*}{Fixed 256-way Zero-shot}
& \multirow{5}{*}{\shortstack[l]{MassBank + custom dataset \\ (Benchmark \cite{bushuiev2024massspecgym})}}
& \multirow{5}{*}{256}
& Random & 0.37 \\
& & & DeepSets & 1.47 \\
& & & Fingerprint FFN & 2.54 \\
& & & DeepSets + Fourier & 5.24 \\
& & & MIST & 14.64 \\

\cmidrule(lr){2-5}
& \multirow{2}{*}{MassBank only}
& \multirow{2}{*}{256}
& Random & 0.39 \\
& & & \textbf{Our method} & \textbf{42.16} \\

\midrule

\multirow{2}{*}{Fixed Global Zero-shot}
& \multirow{2}{*}{MassBank only}
& \multirow{2}{*}{26k}
& Random & $<0.01$ \\
& & & \textbf{Our method} & \textbf{3.56} \\

\bottomrule
\end{tabular}
}
\caption{ Zero-shot molecular retrieval performance measured by Hit@1 (Top-1 accuracy). Benchmark results are reported from MassSpecGym, while our evaluations are conducted on MassBank-only candidate pools with strict scaffold disjointness. }
\label{conditional_zeroshot_comparison}
\end{table*}


\subsubsection{Evaluation Protocols} 
We assess the model under two distinct settings:
\begin{itemize}
    \item \textbf{Zero-Shot Retrieval:} Ranking the true molecule from a large candidate pool under a global setting with all test-set molecules and a 256-candidate benchmark setting for comparison with prior work \cite{bushuiev2024massspecgym}.
    
    \item \textbf{Few-Shot Classification:} 1-way, 5-way classification to evaluate robustness against instrument variability compared to meta-learning baselines \cite{cao2022learning}.
\end{itemize}
We use 600 episodes to obtain statistically stable estimates of both mean accuracy and variance, which is essential for evaluating robustness under episodic settings.
Zero-shot retrieval evaluates whether the aligned embedding space can identify unseen molecules without any task-specific adaptation. Few-shot classification further tests whether the same embedding space remains stable under instrument-induced variability when only a few labeled examples are provided per class.




\subsection{Zero-Shot Molecular Retrieval Results}
We assess the ability of the proposed model to identify correct molecular structures without task-specific fine-tuning.
To contextualize our evaluation, we compare our results with MassSpecGym, a recent benchmark for evaluating zero-shot molecular identification under controlled search protocols.

In Table~\ref{conditional_zeroshot_comparison}, we report zero-shot molecular retrieval performance under fixed candidate settings using Hit@1 as Top-1 accuracy.
The benchmark results are taken from MassSpecGym~\cite{bushuiev2024massspecgym}, which evaluates zero-shot retrieval under a fixed 256-way candidate pool constructed from MassBank together with additional custom datasets.
In this fixed 256-way protocol, each query spectrum is evaluated against a predefined candidate set of 256 molecules, where exactly one molecule is correct.
No precursor-mass information or auxiliary filtering is provided, and the model must identify the correct molecule solely based on the MS spectrum.
Random corresponds to uniform selection within the fixed candidate pool. DeepSets and Fingerprint FFN represent supervised spectrum-to-structure retrieval baselines using set-based aggregation and fingerprint regression, respectively, while DeepSets+Fourier augments DeepSets with Fourier features for improved peak resolution modeling. MIST is a recent strong MS representation model under the same protocol \cite{bushuiev2024massspecgym}.

\subsubsection{Comparison with Benchmark}
While MassSpecGym evaluates zero-shot performance using a candidate pool constructed from MassBank together with additional custom datasets, our evaluation is conducted using MassBank only \cite{bushuiev2024massspecgym}.
Although the exact composition of the additional custom datasets differs from our setting, the Random baseline achieves Hit@1 scores of 0.37 and 0.39 under the two candidate constructions, indicating that dataset composition alone does not substantially affect evaluation difficulty.
Therefore, we can make meaningful comparisons of our methods with benchmark references under evaluation protocols as shown in Table~\ref{conditional_zeroshot_comparison}.

In these evaluations, our model achieves 42.16\% Hit@1, significantly outperforming all reported benchmark baselines.
The results show that effective semantic alignment between MS spectra and molecular representations enables robust zero-shot generalization without relying on hints about mass or task-specific fine-tuning.

\subsubsection{Global Retrieval Performance.}
In addition to the fixed 256-way setting used for comparison with state-of-the-art benchmarks, we further evaluate our method under a substantially more challenging fixed global candidate pool consisting of approximately 26,000 molecules from MassBank.
While the input condition remains identical, this setting increases the candidate space by over two orders of magnitude, requiring the model to select the correct molecule from a much larger set in a zero-shot manner.
Such a large-scale evaluation demands high discriminative precision, as successful retrieval relies on globally well-aligned chemical representations across the entire embedding space.
To the best of our knowledge, zero-shot molecular retrieval under a fixed candidate pool of this scale has not been previously reported.

Even under this extreme condition, our model achieves a Hit@1 of 3.56\% and a Hit@10 of 17.58\%, as shown in Table~\ref{conditional_zeroshot_comparison}.
These results indicate non-trivial large-scale retrieval capability and highlight the potential of the proposed framework for hint-free molecular identification beyond conventional benchmark settings.


\begin{table}[t]
    \centering
    \small
    \resizebox{0.5\textwidth}{!}{
    \begin{tabular}{lcc}
        \toprule
        \textbf{Model} 
        & \textbf{1-shot Accuracy (\%)} 
        & \textbf{5-shot Accuracy (\%)} \\
        \midrule
        Baseline++ (CNN)    & 63.47 & 79.09 \\
        ProtoNet (CNN)     & 62.61 & 78.01 \\
        TSL\_MFL (CNN)     & 64.37 & 73.25 \\
        RelationNet (CNN)  & 52.75 & 69.40 \\
        MatchingNet (FC)   & 59.72 & 67.89 \\
        \midrule
        \textbf{Our Work} 
        & \textbf{88.01} 
        & \textbf{95.40} \\
        \bottomrule
    \end{tabular}
    }
    \caption{
    Average accuracy under 5-way 5-query episodic evaluation over 600 episodes.
    }
    \label{fewshot_accuracy}
\end{table}

\begin{figure*}
    \centering
    \begin{subfigure}[t]{0.47\textwidth}
        \centering
        \includegraphics[width=\linewidth]{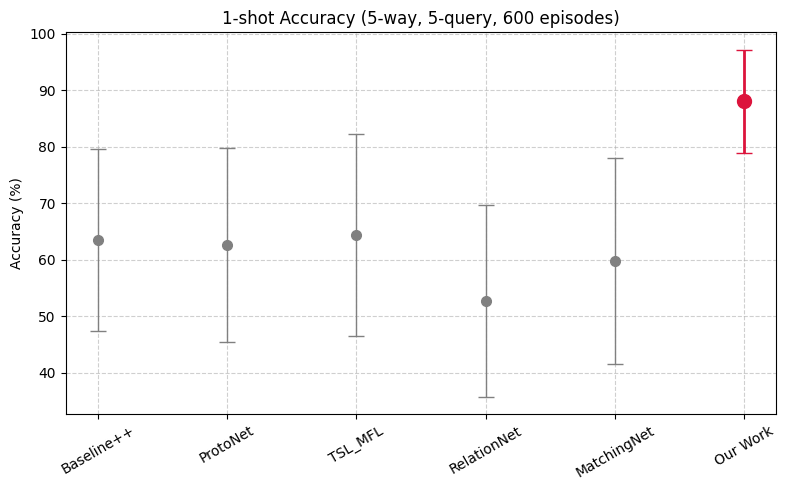}
        \caption{1-shot accuracy (5-way, 5-query, 600 episodes).}
        \label{fig:1shot_errorbar}
    \end{subfigure}
    \hfill
    \begin{subfigure}[t]{0.47\textwidth}
        \centering
        \includegraphics[width=\linewidth]{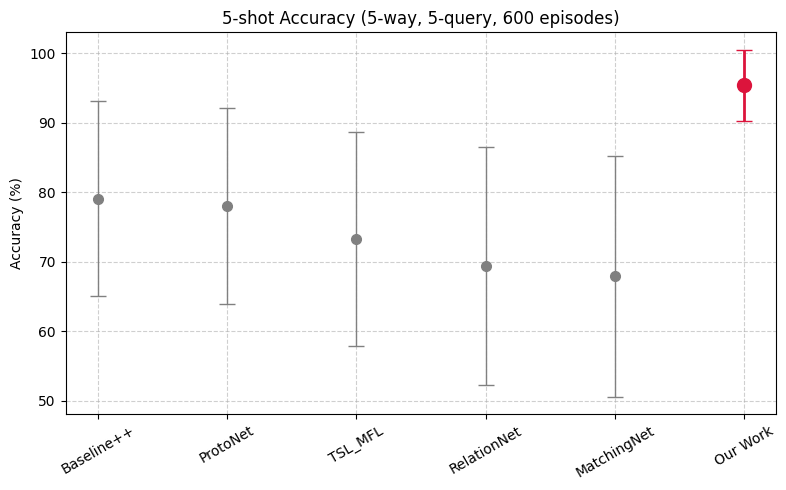}
        \caption{5-shot accuracy (5-way, 5-query, 600 episodes).}
        \label{fig:5shot_errorbar}
    \end{subfigure}

    \caption{
    Few-shot identification performance with episodic variance.
    Error bars indicate standard deviation across evaluation episodes.
    }
    \label{fewshot_errorbar}
\end{figure*}

\subsection{Few-Shot Classification and Comparison}
To validate the effectiveness of our proposed framework under heterogeneous instrument conditions, we conducted 5-way $K$-shot classification experiments.
We compare our method with standard metric-based meta-learning baselines, including TSL-MFL, MatchingNet, RelationNet, and Baseline \cite{sun2025tst_mfl,wei2022erp,zhuang2018relationnet,chen2021meta}.

\subsubsection{Evaluation with Comparative Models.}
As summarized in Table~\ref{fewshot_accuracy}, our framework consistently outperforms all baseline methods in both 1-shot and 5-shot scenarios.
While traditional meta-learning approaches struggle to cope with the domain shift induced by unseen chemical scaffolds and heterogeneous acquisition conditions, our method demonstrates markedly superior generalization.
Specifically, our approach achieves \textbf{88.01\%} accuracy in the 1-shot setting and \textbf{95.40\%} accuracy in the 5-shot setting, suggesting that aligning spectral representations with chemically meaningful semantics yields a more robust molecular structure embedding space than learning purely from spectral statistics.


\subsubsection{Evaluation of Noise Robustness.}
The substantial performance improvement from 1-shot to 5-shot with $\Delta +7.39\%$ suggests that the learned metric space supports effective prototype averaging, which helps suppress spectral variability and measurement noise under previously unseen conditions.
As shown in Fig.~\ref{fewshot_errorbar}, the proposed method achieves both higher mean accuracy and substantially lower episodic variance than metric-based baselines, demonstrating improved stability under heterogeneous instrument conditions.

\begin{table}[t]
\centering
\resizebox{0.49\textwidth}{!}{
\begin{tabular}{lcc}
\toprule
\textbf{Ablation Setting} 
& \textbf{Zero-Shot (R@1, 256-way)} 
& \textbf{5-way 5-shot Acc. (\%)} \\
\midrule
Full Model (Ours) 
& 42.16 
& 95.40 \\
Without Gaussian Fourier Projection
& 32.95 
& 89.44 \\
Without Contrastive Loss (MSE)
& 38.54 
& 91.59 \\
Frozen ChemBERTa Encoder
& 40.55 
& 95.32 \\
\bottomrule
\end{tabular}
}
\caption{
Ablation study on components of the proposed framework.
}
\label{ablation_summary}
\end{table}
\subsection{Ablation Study}

Table~\ref{ablation_summary} summarizes the ablation results of the proposed framework.
Removing the Gaussian Fourier projection consistently degrades performance in both zero-shot retrieval and few-shot classification, demonstrating that preserving fine-grained mass-to-charge variations is critical for robust generalization under heterogeneous instrument conditions.

Replacing the contrastive InfoNCE objective with a mean squared error loss results in a noticeable performance drop, indicating that contrastive alignment is essential for effective cross-modal generalization between MS spectra and molecular embeddings.
Freezing the pretrained ChemBERTa encoder results in only a marginal performance change, indicating that the proposed cross-modal alignment effectively leverages the semantic structure already encoded in the chemical language model, rather than relying on task-specific fine-tuning.

Overall, these results show that the performance gains of the proposed framework do not stem from isolated architectural choices, but from a principled integration of physical spectral modeling and molecular structure embeddings, enabling stable and transferable molecular identification across heterogeneous MS instruments.

\section{Conclusion}\label{sec5}

We proposed a contrastive cross-modal transfer learning framework for molecular identification across heterogeneous MS instruments.
Through explicit alignment between MS spectra and molecular structure embeddings, the proposed framework acquires domain-invariant representations that generalize effectively across unseen molecular scaffolds and novel instruments.
This alignment enables the model to bridge spectral signal patterns with chemically meaningful molecular structure embeddings, rather than relying on instrument-specific spectral statistics.
Experimental results demonstrate consistent performance gains in both zero-shot and few-shot settings, outperforming existing metric-based and cross-domain learning approaches while exhibiting substantially improved stability under instrument-level domain shifts.

Future work will explore extending the proposed framework to additional sensing modalities and further strengthening cross-domain generalization in heterogeneous sensing environments.

\appendix





\section*{Acknowledgments}
This work was supported by Korea Research Institute of defense Technology planning and advancement (KRIT) grant funded by Defense Acquisition Program Administration (DAPA) (KRIT-CT-21-034).

\newpage
\bibliographystyle{named}
\bibliography{ijcai25}

\end{document}